\documentclass[conference,a4paper]{IEEEtran}
\IEEEoverridecommandlockouts

\usepackage[hidelinks]{hyperref}
\usepackage[cmex10]{amsmath}
\usepackage{amssymb,amsfonts}
\usepackage{dblfloatfix}

\usepackage[ruled,vlined]{algorithm2e}
\usepackage{graphicx}
\graphicspath{{Figures/PDF/}{Figures/PNG/}}

\usepackage{booktabs}
\usepackage{siunitx}
\usepackage[numbers,compress]{natbib}
\usepackage{texnames}
\usepackage{bm,bbm}
\usepackage{orcidlink}

\begin{document}

\title{\uppercase{Deep Learning Super Resolution For Satellite Cloud Mask downscaling}
}

\author{
\IEEEauthorblockN{
Angelos~Georgakis\IEEEauthorrefmark{1},
Valentina~Kanaki\IEEEauthorrefmark{1},
Giorgos~Giannopoulos\IEEEauthorrefmark{1},
Stella~Girtsou\IEEEauthorrefmark{1},\\
Ioannis~Kontogiorgakis\IEEEauthorrefmark{1},
Charalampos~Kontoes\IEEEauthorrefmark{1},
Kostas~Philippopoulos\IEEEauthorrefmark{2}
}
\IEEEauthorblockA{\IEEEauthorrefmark{1}
BEYOND EO Centre, IAASARS, National Observatory of Athens, Athens, Greece\\
\{ageorgakis, valekan, giannopoulos, sgirtsou, ikontog, kontoes\}@noa.gr
}
\IEEEauthorblockA{\IEEEauthorrefmark{2}
National and Kapodistrian University of Athens, Faculty of Physics,
Section of Environmental Physics and Meteorology,\\ Athens, Greece\\
kphilip@phys.uoa.gr
}
}

\maketitle
\begin{abstract}
A vast amount of optical satellite data is being transmitted to Earth-based servers every day, and more than half of this data is affected by haze or clouds. Additionally, this data suffers from the fundamental trade-off between spatial and temporal resolution, which remains largely unresolved, making the acquisition of continuous high-resolution satellite observations of clouds an ongoing challenge. This work addresses this challenge by proposing two Deep Learning super-resolution methods for the accurate downscaling of SEVIRI cloud mask products, as well as a novel cross-sensor cloud mask dataset called SEVMOD-CM, created by spatially and temporally matching MODIS and SEVIRI satellite observations.
The two proposed models are a CNN-based (SpatialCNN) and a GAN-based (SpatialGAN) Neural Network. Trained on the SEVIRI spectral and cloud mask products, the proposed methods predict the corresponding MODIS Cloud masks, achieving a 4× spatial enhancement across sensor domains. Both approaches are evaluated experimentally, and compared against the standard bicubic interpolation upsampling technique. The experimental results demonstrate the value of the proposed models and dataset for the remote sensing community, highlighting the benefits of applying super-resolution techniques to geostationary-derived cloud mask products for applications such as atmospheric monitoring, weather forecasting, disaster risk reduction, solar energy forecasting, and climate research.

\end{abstract}


\section{Introduction} 

\noindent Over the past decades remote sensing (RS) imagery has emerged as a central field of research, 
with crucial implications in applications such as Earth observation (EO) and environmental monitoring. Since
\~ 67\% of the globe is covered by haze or clouds \cite{6422379}, optical RS retrievals are consequently greatly affected due to the scattering and attenuation of the electromagnetic signals leading to a deficiency of surface information \cite{9956865}. Satellite retrievals of cloud formations that are of both high temporal and spatial resolution, are crucial for many application such as solar energy forecasting, weather prediction and disaster management, as many of such application utilize the cloud mask (CM) products derived from Geostationaty satellites such as MSG. However, due to technical limitations, such as sensor size, orbital altitude, and data transmission bandwidth, 
the persistent trade-off between spatial and temporal resolution (no sensor can capture information both at the highest possible spatial and temporal resolution across all wavebands) remains a significant challenge in the field of RS, that greatly affects all the CM dependent applications. Furthermore, cloud misclassifications reported in existing cloud mask products  \cite{Letu:14}, \cite{Leinenkugel20042013}, together with the limited availability of high-resolution imagery for cloud removal (CR) applications \cite{10354035}, further stretch the need for accurate high-resolution CM products. Obtaining CM products of high temporal and spatial resolution has also been emphasized in studies on solar radiation nowcasting and forecasting \cite{amt-17-1851-2024}, and on cloud-gap filling for applications such as grassland mowing event detection \cite{TSARDANIDIS2025109732} among others.
Super-resolution (SR) is a well established set of algorithmic techniques in the field of computer vision and image processing that aim to improve the visual analysis of images. 
The term Super-Resolution is often used as an umbrella term to include many image upsampling techniques that have been widely categorized as, Interpolation-based \cite{6161647}, \cite{8336891}, Reconstruction-based \cite{YANG2018163}, \cite{8421656} and Learning-based \cite{Liu2019} \cite{8723565} algorithms.
A widely adopted interpolation method is Bicubic interpolation \cite{1163711}, as it is a simple and explainable type of image upsampling that produces smooth HR images, however, it is  known to suffer from the chessboard effect \cite{4270053}.
Reconstruction-based SR algorithms on the other hand, degrade rapidly when given large magnification factors or relatively few input images. In these cases, the result may be lacking important high-frequency details \cite{1033210}. Therefore the emphasis is being given in the advancement of DL SR algorithms. Various such approaches have been proposed, which can be roughly categorized in the following two groups: Convolutional Neural Network (CNN)-based and Generative Adversarial Network (GAN)-based methods, each one with its own advantages and disadvantages \cite{https://doi.org/10.1049/ipr2.12760}. The CNN-based methods originate from the SRCNN paper \cite{7115171}, where a 3-layer CNN was proposed to perform SR on all 3 color channels simultaneously, surpassing the then SotA methods. Soon after a GAN-based network for image SR was proposed \cite{ledig2017photorealisticsingleimagesuperresolution}, (SRGAN), obtaining better overall structural similarity. Since then, these models have been widely adjusted to adapt to different SR RS tasks \cite{rs13061104}, \cite{7937881}, but continue to provide the fundamental approaches on SR \cite{rs14215423}.
Despite the widespread application of SR in different RS tasks, the cross-sensor CM SR task is one that remains largely unexplored. In this paper we introduce two advanced DL models for cloud mask SR, as well as a novel DL-ready cross-sensor CM dataset, called SEVMOD-CM, and demonstrate its applicability. The proposed models are drawn from the two major SR DL families: a CNN-based model (SpatialCNN) and a GAN-based model (SpatialGAN). In both architectures, we implement task specific extensions and enhancements (detailed in Section \ref{SR_models}) to enable effective learning that derive from the key characteristics of the CM products they are trained on. The models are implemented and evaluated against the standard bicubic interpolation upsampling method as a baseline. For the dataset creation, data from satellite sensors which differ in spatial, spectral, and temporal resolutions are combined. This data is derived from: (a) the MODIS sensor onboard the Aqua and Terra satellites, used as the High-Resolution (HR) ground truth for model training and (b) the SEVIRI instrument aboard the MSG satellite, serving as the Low-Resolution (LR) input to the SR models. These models are expected to significantly enhance the ability to detect and monitor cloud formations in near real time, thereby supporting faster responses to extreme weather events and mitigating their potential impacts. Moreover, improved CM spatial resolution can contribute to more accurate solar radiation monitoring and forecasting, facilitating its more effective integration into the power grid. To the best of the authors’ knowledge, this is the first effort to apply DL SR methods directly to the CM products of different satellite sensors.\\
The key contributions of this study are summarized as follows.
\begin{itemize}   
    \item we implement two advanced SR algorithms in order to improve the spatial analysis of the Cloud Mask product generated by the SEVIRI sensor aboard the MSG geostationary satellite
    \item we create SEVMOD-CM, a comprehensive cross sensor cloud mask oriented, AI-ready dataset, and demonstrate its usability
    \item we evaluate the proposed algorithms, and explore limitations and possible future research directions
\end{itemize}

\section{MATERIALS AND METHODS}
\noindent Regarding the terminology used in this paper, as is usually the case in EO and RS research, the term “downscale” refers to the transition from low to high resolution, indicating the decrease of the actual spatial trace of a pixel on the map, while in computer vision studies the same process is described by the word "upscale", referring to the increase in image analysis 
with "downscale" indicating the opposite process. The words "upsampling" and "downsampling" are also used instead of upscaling/downscaling. 
In this paper the term “upsample” is being prevalently used.\\
In this section, the design choices and implementation steps that were carried out are analyzed. Specifically the four basic steps below will be addressed:
\begin{itemize}
    \item Finding and downloading the necessary data
    \item Data postprocessing and SEVMOD-CM creation
    \item Selection and implementation of SR DL models
    \item SR result evaluation
\end{itemize}
These stages are presented in more detail below, while a clear roadmap of the dataset creation methodology is presented 

\subsection{SEVMOD-CM DATASET}
\subsubsection{DATA ACQUISITION}
For the development of the SEVMOD-CM dataset, paired low- and high-resolution satellite observations were collected from the SEVIRI instrument onboard Meteosat Second Generation (MSG) and the MODIS instruments onboard Aqua and Terra. Specifically, the Seviri 11 spectral channels and the CM product of 3km spatial resolution, which was separately downloaded but later processed in a combined netcdf file, as well as the Modis CM product of 1km spatial resolution, spanning 2020–2023 and covering Europe and North Africa. MODIS CM products (MOD35\_L2 for Terra and MYD35\_L2 for Aqua) were used as the high-resolution reference, and their acquisition times were used to define the SEVIRI candidate timestamps, within a time window of 15 minutes, to account for the much higher temporal sampling of MSG. MODIS data were retrieved from NASA Earthdata \footnote{ \href{https://www.earthdata.nasa.gov/data/catalog}{https://www.earthdata.nasa.gov/data/catalog}.} and SEVIRI data from EUMETSAT \footnote{ \href{https://data.eumetsat.int/search?query=}{https://data.eumetsat.int/search?query=}}, resulting in a dataset of approximately 200 GB stored in a local server.

\subsubsection{DATA PROCESSING}
The downloaded MODIS and SEVIRI products were preprocessed to generate paired image patches, that are temporally synchronized and spatially aligned, and suitable for the training of DL SR algorithms. The MODIS CM product initially provides a large set of ancillary flags and test results encoded in 48-bit representation. By applying bitwise operators it was decoded and converted into a binary cloud mask, while the SEVIRI CM that initially contained categorical values (0: clear sky over water, 1: clear sky over land, 2: clouds, 3: raw or no data points, i.e. areas outside the Earth’s disk) indicating cloud conditions, was similarly binarized and appended as a twelfth SEVIRI channel. SEVIRI products, originally provided in a geostationary projection, were reprojected to EPSG:4326 to match the MODIS Coordinate Reference System (CRS) and allow for precise geolocation and spatial overlapping tests. Spatially overlapping regions were then identified and fixed 128 × 128 MODIS patches were extracted. Thus spatially corresponding SEVIRI patches were cropped, but not initially forced on a specific image size, to allow for covering the exactly same geographic region as MODIS. The cropped SEVIRI patches were then resampled to a canonical 32 × 32 grid, and paired with the MODIS ones. Invalid or empty patch pairs were discarded. The resulting paired patches were stored as GeoTIFF files with full metadata and split into training, validation, and test sets using an 80/10/10 ratio.
\begin{figure}[h]
    \centering
        \includegraphics[width=\linewidth]{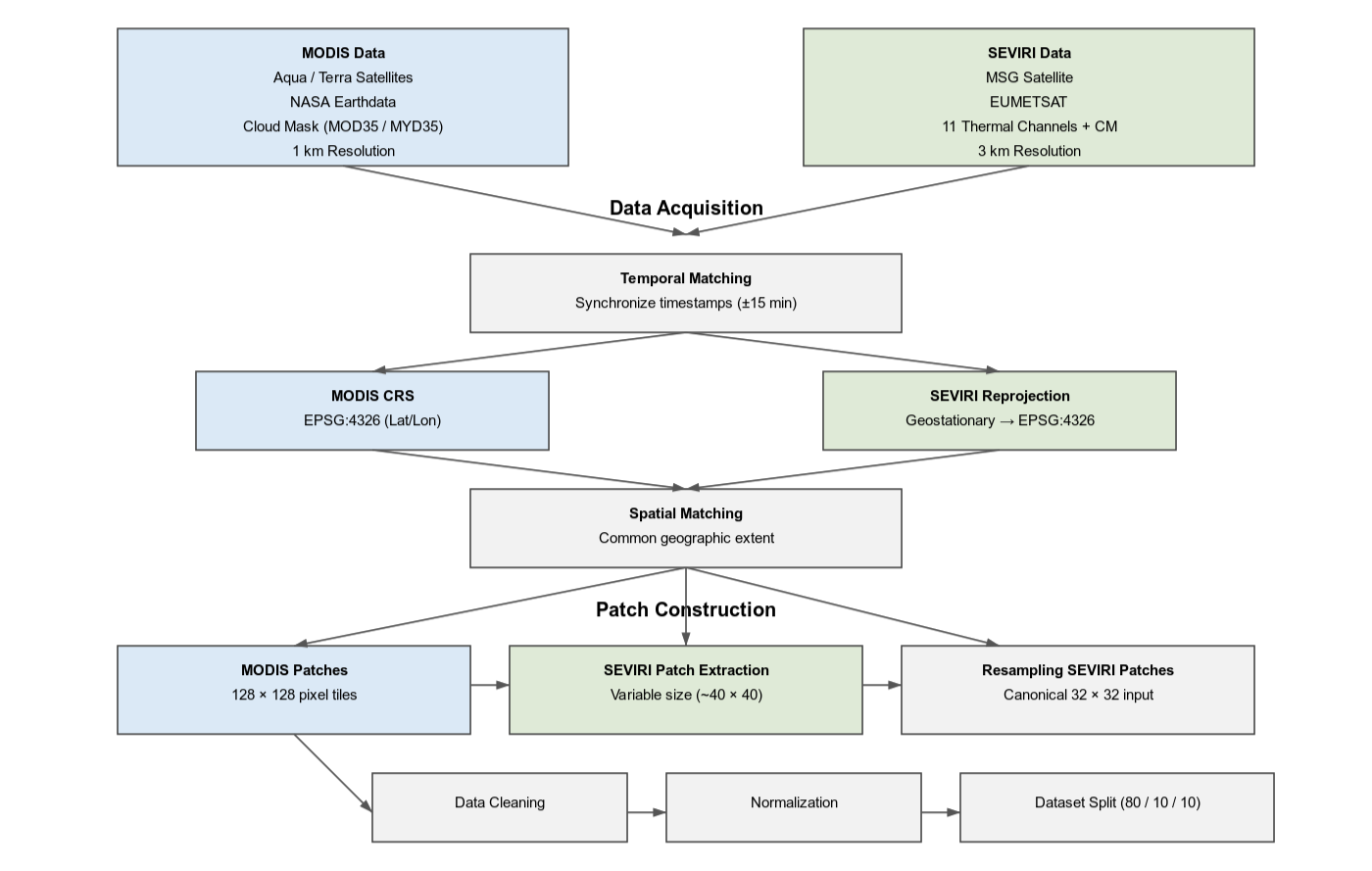}
        \caption{SEVMOD-CM Dataset creation methodology}
        \label{fig:SEVMOD-CM dataset methodology} 
\end{figure}

    

\subsection{Super Resolution Deep Learning Models}
\label{SR_models}
\begin{figure}[h]
    \centering
        \includegraphics[width=\linewidth]{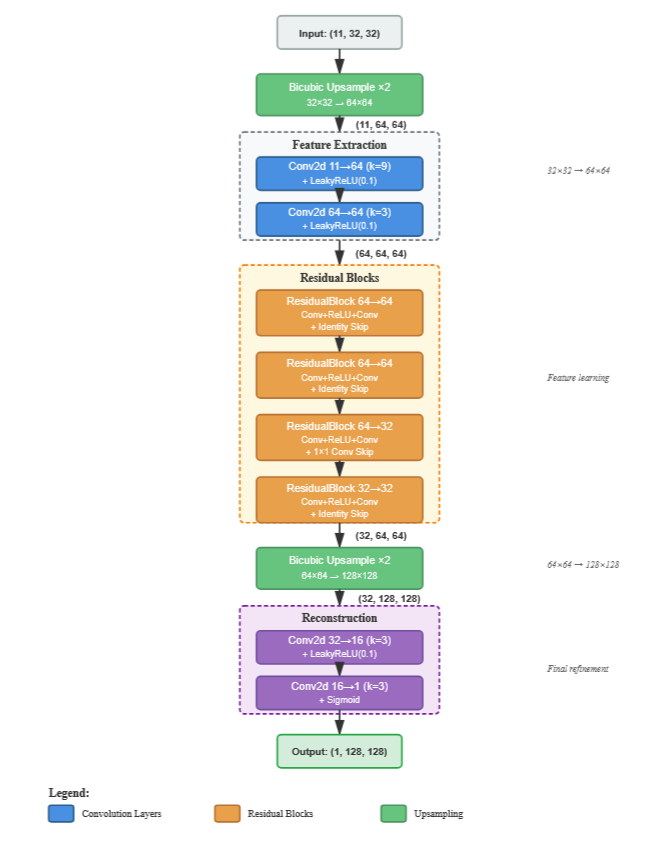}
        \caption{SpatialCNN architecture}
        \label{fig:spatialcnn_arch} 
\end{figure}

For the task of SR, we developed two distinct DL architectures to specifically address cross-sensor and cross-resolution discrepancies between geostationary SEVIRI and polar-orbiting MODIS observations. The task was formulated as a computer vision supervised problem, where we mapped LR multi-spectral SEVIRI radiances and the associated cloud masks, to the spatially and temporally corresponding HR MODIS cloud masks. 
The first model is SpatialCNN, depicted in Fig. \ref{fig:spatialcnn_arch}. It is a deep CNN model that extends traditional CNN SR approaches \cite{7115171} by incorporating residual learning and progressive upsampling. These architectural configurations are based on thorough RS SR domain knowledge, and are proven to enhance image reconstruction fidelity \cite{10.1007/978-3-030-67070-2_2}, \cite{9329109}, \cite{9416983}, \cite{9645163}, \cite{7780576}, \cite{10236467}, \cite{Wang_2018_CVPR_Workshops}. On top of that, in order to adjust the model to the task of binary CM SR, the proposed architecture also supports a flexible number of input channels, allowing the network to ingest between one and twelve SEVIRI bands, or any arbitrary subset in between, thus enabling the systematic evaluation of spectral contribution to CM SR. Finally to adjust to the binary nature of the MODIS target cloud mask, BCE was the chosen loss function, while a sigmoid-bounded single-channel output constrains predictions to probabilistic CM values.

\begin{figure}[h]
    \centering
        \includegraphics[width=\linewidth]{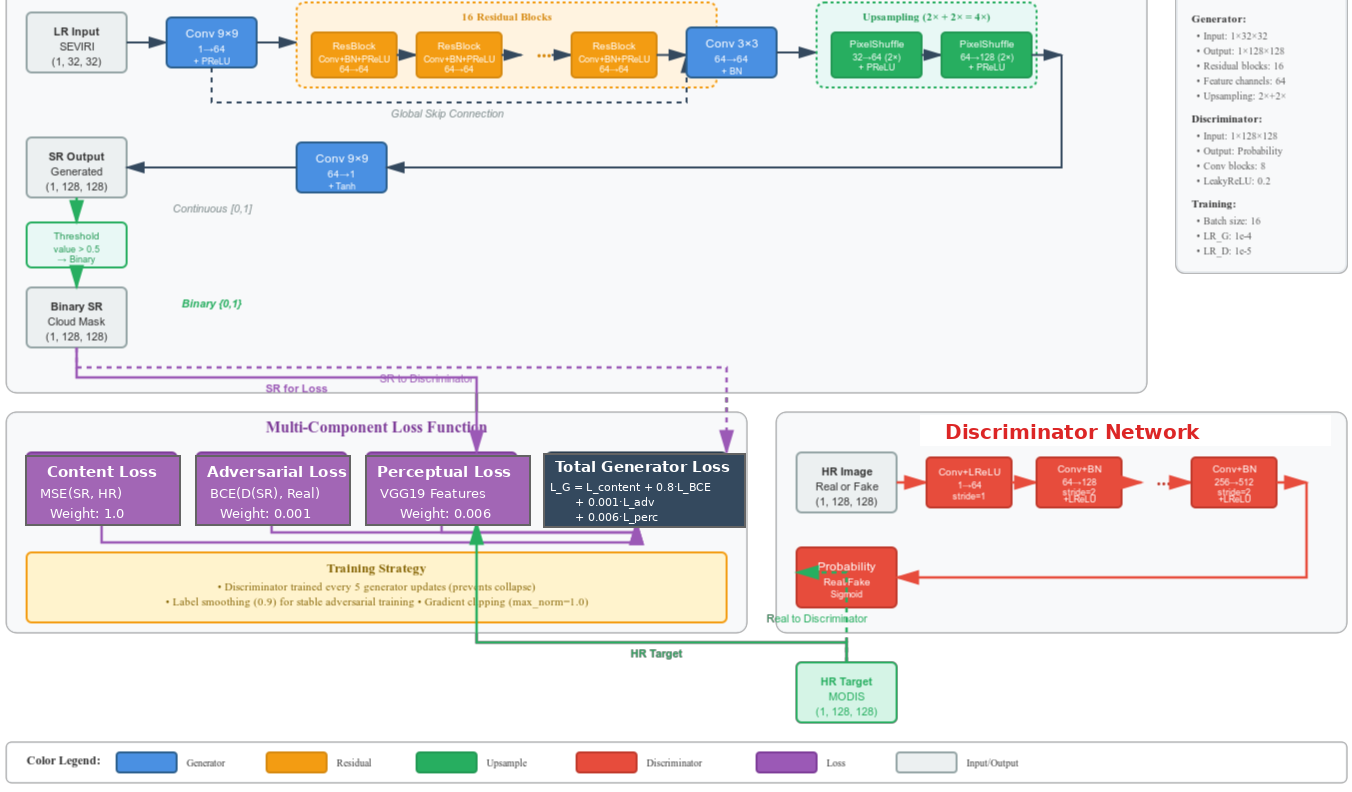}
        \caption{SpatialGAN architecture}
        \label{fig:spatialgan_arch} 
\end{figure}

The second model, SpatialGAN \ref{fig:spatialgan_arch}, is a GAN model based on SRGAN \cite{ledig2017photorealisticsingleimagesuperresolution}, but specifically adjusted to the binary CM SR task. 
While also modified to work with a flexible number of SEVIRI input channels, the key modification comes in the composite loss function of SpatialGAN. The SRGAN framework was originally proposed for photo-realistic image reconstruction, thus its loss formulation required reinterpretation for binary cloud mask SR. We extended it by adding a BCE term to explicitly enforce pixel-wise cloud classification consistency and faster convergence, while the initial perceptual and adversarial losses were kept but downgraded by adding a weight factor, to act as regularizers that encourage spatial coherence.\\
Model performance was evaluated on a held-out test set using Peak Signal-to-Noise Ratio (PSNR), Mean Squared Error (MSE), and Structural Similarity Index (SSIM), which were selected to quantify the reconstruction error and the spatial consistency of the unavoidable cross-sensor image misalignment. 

\section{RESULTS AND DISCUSSION}
\noindent\textbf{Dataset.} The present work resulted in the creation of the SEVMOD-CM Dataset, consisting of 4.8GB of data, with a total of  126216 \textit{.tif} files, meaning 63108 complete MODIS-SEVIRI pairs. All the SEVIRI patches, contained 12 bands, 11 spectral and the binarized Cloud Mask product, while the MODIS, serving as the model target, has just the binary CM created during the preprocessing phase. Of these initial 63108 pairs, after filtering out the ones that contained no clouds at all 20828 final paired patches (33\% retention rate) were kept to feed into the networks.






\noindent\textbf{Evaluation Setting.}
The implementation of our models was done using the PyTorch framework. Initial experimentation on the training dataset lead us to the following hyperparameter configurations. Regarding SpatialCNN, the set of SEVIRI input channels that performed best is: 2,4,6,12.
The network was trained for 29 epochs as early-stopping was implemented. LeakyReLU was chosen as the activation function of the hidden layers, while for the output layer the sigmoid function was used to constrain outputs to [0,1]. Adam was the selected optimizer algorithm, while a Batch size of 8 was used. During inference a threshold of 0.5 was used in order to retrieve the cloud masks in binary form, thus comparable to the SEVIRI provided ones. \\
Regarding the SpatialGAN model, we trained the network for 100 epochs using SEVIRI channels 2, 4, 6, and 12 as input as well, with a batch size of 16. Both the generator and discriminator were optimized using the Adam optimizer with learning rates of $1 \times 10^{-4}$. Finally to reduce discriminator overconfidence and stabilize the adversarial training process, the discriminator was updated less often than the generator with a 5:1 ratio, and a label smoothing of 0.9 was applied, helping prevent discriminator dominance. 
As for the baseline bicubic interpolation method, it was applied using PyTorch’s default kernel parameters (Keys’ cubic convolution kernel with a=-0.75)\footnote{ \href{https://github.com/pytorch/pytorch/tree/main}{https://github.com/pytorch/pytorch/tree/main}}.\\
\noindent\textbf{Evaluation Results.} The experimental findings in Table ~\ref{tab:sr_metrics} demonstrate the comparative performance of the applied SR models on the held-out test set of the SEVMOD-CM dataset. As shown in Table~\ref{tab:sr_metrics}, SpatialGAN achieved by far the highest SSIM, as well as the lowest MSE values, while its PSNR was marginally lower compared to the other methods. On the other hand SpatialCNN performed similarly but slightly worse than bicubic inerpolation across all 3 evaluation metrics, surpassing only SpatialGAN in PSNR.

\begin{table}[t]
\centering
\caption{Quantitative evaluation of super-resolution methods on the SEVMOD-CM test set.}
\label{tab:sr_metrics}
\begin{tabular}{lccc}
\hline
\textbf{Metric} 
& \textbf{SpatialCNN} 
& \textbf{SpatialGAN} 
& \textbf{Bicubic Interpolation} \\
\hline
PSNR (dB) & 17.99 & 17.9 & 18.32 \\
SSIM      & 0.534 & 0.7811 & 0.578 \\
MSE       & 0.173 & 0.0480 & 0.12 \\
\hline
\end{tabular}
\end{table}

\begin{figure}[h]
    \centering
    \includegraphics[width=1\linewidth]{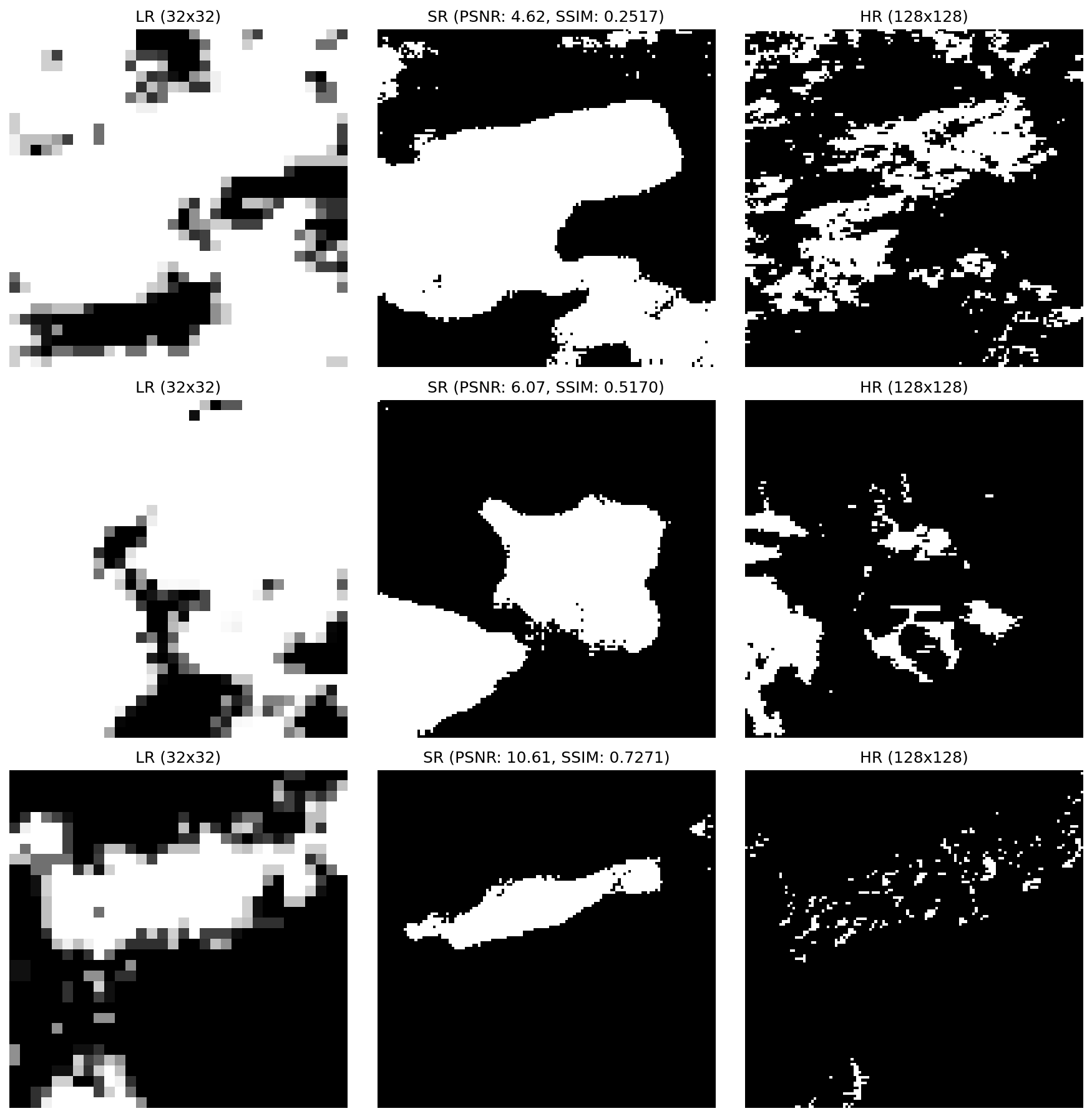}
    \caption{Test results of SpatialGAN super-resolved images (middle), alongside the initial SEVIRI CM (left) and the target MODIS CM (right)}
    \label{fig:srgan_test}
\end{figure}

Overall, SpatialGAN succeeded in producing sharper and more perceptually accurate reconstructions than SpatialCNN, while when compared to the bicubic interpolation baseline approach it achieved slighlty lower PSNR but significantly higher SSIM and lower MSE. This indicates that while it might have missed some high frequency details, the overall structural similarity of the reconstructed cloud masks was greatly improved, highlighting the advantages of adversarial training for SR of satellite imagery. This is evident in figure \ref{fig:srgan_test}, where the clear structural enhancement of the  SEVIRI cloud shapes is visually demonstrated. While the MODIS targets' complex and scattered cloud patterns result in lower numerical metrics, we see that SpatialGAN has successfully learned to reconstruct the overall cloud structure and distribution, demonstrating the visual fidelity and perceptual quality of the model's outcomes. \\ 
Regarding the limitations of the method, the inherent spatial, temporal and spectral misalignment in cross-sensor imagery, remains a fundamental challenge that image upsampling methods cannot yet fully resolve. On the other hand, future research will investigate advanced DL architectures, including Vision Transformers and diffusion models to better address the cross-sensor SR challenge, as well as applying them to different cloud related products. 


\section{CONCLUSION}
\noindent This paper presented a novel approach for cross-sensor satellite CM SR, using SEVIRI and MODIS satellite imagery and DL architectures. We developed two distinct advanced DL SR models, as well as a task specific dataset, and showcased their applicability. Specific model configurations exhibited much higher structural similarity and reconstruction accuracy compared to the baseline method, highlighting the effectiveness of GAN-based models in such tasks. These findings emphasize on the potential of SR methods utilizing geostationary observations as a valuable tool for near real-time cloud monitoring.
\small
\bibliographystyle{IEEEtranN}
\bibliography{references}

\end{document}